# Multiclass Approaches for Support Vector Machine Based Land Cover Classification

Mahesh Pal

Lecturer, Department of Civil engineering

National Institute of Technology

Kurukshetra, 136119, Haryana (INDIA)

mpce_pal@yahoo.co.uk

**Abstract:** SVMs were initially developed to perform binary classification; though, applications of binary classification are very limited. Most of the practical applications involve multiclass classification, especially in remote sensing land cover classification. A number of methods have been proposed to implement SVMs to produce multiclass classification. A number of methods to generate multiclass SVMs from binary SVMs have been proposed by researchers and is still a continuing research topic. This paper compares the performance of six multi-class approaches to solve classification problem with remote sensing data in term of classification accuracy and computational cost. *One vs. one, one vs. rest, Directed Acyclic Graph (DAG), and Error Corrected Output Coding (ECOC)* based multiclass approaches creates many binary classifiers and combines their results to determine the class label of a test pixel. Another catogery of multi class approach modify the binary class objective function and allows simultaneous computation of multiclass classification by solving a single optimisation problem. Results from this study conclude the usefulness of *One vs. One* multi class approach in term of accuracy and computational cost over other multi class approaches.

# 1. INTRODUCTION

A new classification system based on statistical learning theory (Vapnik, 1995), called the support vector machine (Boser *et al.*, 1992) has recently been applied to the problem of remote sensing data classification (Foody and Mathur, 2004; Gualtieri and Cromp, 1998; Huang et al., 2002; Pal and Mather, 2003; Zhu and Blumberg, 2002). This technique is said to be independent of the dimensionality of feature space as the main idea behind this classification technique is to separate the classes with a surface that maximise the margin between them, using boundary pixels to create the decision surface. The data points that are closest to the hyperplane are termed "support vectors". The number of support vectors is thus small as they are points close to the class boundaries (Vapnik, 1995). One major advantage of support vector classifiers is the use of quadratic programming, which provides global minima only. The absence of local minima is a significant difference from the neural network classifiers. Like neural classifiers, applications of SVMs to any classification problem require the determination of several user-defined parameters. Some of these parameters are the choice of a suitable multiclass approach, Choice of an appropriate kernel and related parameters, determination of a suitable value of regularisation parameter (i.e. C) and a suitable optimisation technique. SVMs were initially developed to perform binary classification; though, applications of binary classification are very limited. Most of the practical applications involve multiclass classification, especially in remote sensing land cover classification. A number of methods have been proposed to implement SVMs to produce multiclass classification. Most of the research in generating multiclass support vector classifiers can be divided in two categories. One approach involves in constructing several binary classifiers and combing their results while other approach considers all data in one optimisation formulation. This paper compares the performance of some of the multi class approaches in term of classification accuracy and the computational cost for land cover classification using remote sensing data.

## 2. SUPPORT VECTOR MACHINES

SVM are based on statistical learning theory and have the aim of determining the location of decision boundaries that produce the optimal separation of classes (Vapnik 1995). In the case of a two-class pattern recognition problem in which the classes are linearly separable the SVM selects from among the infinite number of linear decision boundaries the one that minimises the generalisation error. Thus, the selected decision boundary will be one that leaves the greatest margin between the two classes, where margin is defined as the sum of the distances to the hyperplane from the closest points of the two classes (Vapnik, 1995). This problem of maximising the margin can be solved using standard Quadratic Programming (QP) optimisation techniques. The data points that are closest to the hyperplane are used to measure the margin; hence these data points are termed 'support vectors'. Consequently, the number of support vectors is small (Vapnik, 1995).

If the two classes are not linearly separable, the SVM tries to find the hyperplane that maximises the margin while, at the same time, minimising a quantity proportional to the number of misclassification errors. The trade-off between margin and misclassification error is controlled by a user-defined constant (Cortes and Vapnik, 1995). SVM can also be extended to handle non-linear decision surfaces. Boser et al.

(1992) propose a method of projecting the input data onto a high-dimensional feature space using kernel functions (Vapnik 1995) and formulating a linear classification problem in that feature space. Further, more detailed discussion of the computational aspects of SVM can be found in Vapnik (1995) and Cristianini and Shawe-Taylor, (2000).

SVM were initially designed for binary (two-class) problems. When dealing with multiple classes, an appropriate multi-class method is needed. Vapnik (1995) suggested comparing one class with the others taken together. This strategy generates $n$ classifiers, where $n$ is the number of classes. The final output is the class that corresponds to the SVM with the largest margin, as defined above. For multi-class problems one has to determine $n$ hyperplanes. Thus, this method requires the solution of $n$ QP optimisation problems, each of which separates one class from the remaining classes. This strategy can be described as 'one against the rest'.

A second approach is to combine several classifiers ('one against one'). Knerr et al. (1990) perform pair-wise comparisons between all $n$ classes. Thus, all possible two-class classifiers are evaluated from the training set of $n$ classes, each classifier being trained on only two out of $n$ classes, giving a total of $n(n-1)/2$ classifiers. Applying each classifier to the test data vectors gives one vote to the winning class. The data is assigned the label of the class with most votes. The results of a recent analysis of multi-class strategies are provided by Hsu and Lin (2002).

**3. SVM for Multiclass Classification**

Originally, SVMs were developed to perform binary classification. However, applications of binary classification are very limited especially in remote sensing land cover classification where most of the classification problems involve more than two classes. A number of methods to generate multiclass SVMs from binary SVMs have been proposed by researchers and is still a continuing research topic. This section provides a brief description of some methods implemented to solve multi-class classification problem with SVM in present study.

**3.1 One against the Rest approach**

This method is also called *winner-take-all* classification. Suppose the dataset is to be classified into $M$ classes. Therefore, $M$ binary SVM classifiers may be created where each classifier is trained to distinguish one class from the remaining $M$-1 classes. For example, class one binary classifier is designed to discriminate between class one data vectors and the data vectors of the remaining classes. Other SVM classifiers are constructed in the same manner. During the testing or application phase, data vectors are classified by finding margin from the linear separating hyperplane. The final output is the class that corresponds to the SVM with the largest margin.

However, if the outputs corresponding to two or more classes are very close to each other, those points are labeled as *unclassified*, and a subjective decision may have to be made by the analyst. Otherwise, a reject decision (Schölkopf and Smola, 2002) may also be applied using a threshold to decide the class label. This multiclass method has an advantage in the sense that the number of binary classifiers to construct equals the number of classes. However, there are some drawbacks. First, during the training phase, the memory requirement is very high and amounts to at the square of the total number of training samples. This may cause problems for large training data sets and may lead to computer memory problems. Second, suppose there are $M$ classes and each has an equal number of training samples. During the training phase, the ratio of training samples of one class to rest of the classes will be $1:(M-1)$. This ratio, therefore, shows that training sample sizes will be unbalanced. Because of these limitations, the *one against one* approach of multiclass classification has been proposed.

### 3.2 One against One Approach

In this method, SVM classifiers for all possible pairs of classes are created (Knerr *et al.*, 1990; Hastie and Tibshirani, 1998). Therefore, for $M$ classes, there will be binary classifiers. The output from each classifier in the form of a class label is obtained. The class label that occurs the most is assigned to that point in the data vector. In case of a tie, a tie-breaking strategy may be adopted. A common tie-breaking strategy is to randomly select one of the class labels that are tied.

The number of classifiers created by this method is generally much larger than the previous method. However, the number of training data vectors required for each classifier is much smaller. The ratio of training data vector size for one class against another is also. Therefore, this method is considered more symmetric than the One-against-the-rest method. Moreover, the memory required to create the kernel matrix is much smaller. However, the main disadvantage of this method is the increase in the number of classifiers as the number of class increases. For example, for 7 classes of interest, 21 classifiers will be created.

### 3.3 Decision Directed Acyclic Graph based Approach

Platt *et al.* (2000) proposed a multiclass classification method called *Directed Acyclic Graph SVM* (DAGSVM) based on the *Decision Directed Acyclic Graph* (DDAG) structure that forms a tree-like structure. The DDAG method in essence is similar to pairwise classification such that, for an *M* class classification problem, the number of binary classifiers is equal to $\frac{1}{2}M(M-1)$ and each classifier is trained to classify two classes of interest. Each classifier is treated as a node in the graph structure. Nodes in DDAG are organized in a triangle with the single root node at the top and increasing thereafter in an increment of one in each layer until the last layer that will have *M* nodes.

The DDAG evaluates an input vector **x** starting at the root node and moves to the next layer based on the output values. For instance, it exits to the left edge if the output from the binary classifier is negative, and it exits to the right edge if the output from the binary classifier is positive. The binary classifier of the next node is then evaluated. The path followed is called the *evaluation path*. The DDAG method basically eliminates one class out from a list. Initially the list contains all classes. Each node evaluates the first class against the last class in the list. For example, the root node evaluates class 1 against class *M*. If the evaluation results in one class out of two classes, the other is eliminated from the list. The process then tests the first and the last class in the new list. It is terminated when only one class remains in the list. The class label associated with the input data will be the class label of the node in the final layer of the evaluation path or the class remained in the list. Although the number of binary classifiers still equals the pairwise classification method, the inputs are evaluated $M-1$ times instead of $\frac{1}{2}M(M-1)$ times as is the case with pairwise classification.

### 3.4 Multiclass Objective Function

Instead of creating many binary classifiers to determine the class labels, this method attempts to directly solve a multiclass problem (Weston and Watkins, 1998, Lee *et al.*, 2001; Crammer and Singer, 2001; Schölkopf and Smola, 2002). This is achieved by modifying the binary class objective function and adding a constraint to it for every class. The modified objective function allows simultaneous computation of multiclass classification and is given by (Weston and Watkins, 1998),

$$\min_{\mathbf{w},b,\xi}\left[\frac{1}{2}\sum_{i=1}^{M}\|w\|^2 + C\sum_{i=1}^{k}\sum_{r\neq y_i}\xi_i^r\right]$$

subject to the constraints,

$$\mathbf{w}_{y_i}\cdot\mathbf{x}_i + b_{y_i} \geq \mathbf{w}_r\cdot\mathbf{x}_i + b_r + 2 - \xi_i^r \quad \text{for}$$

and,

$$\xi_i^r \geq 0 \qquad \text{for } i=1,...,k$$

where $y_i \in \{1,...,M\}$ are the multiclass labels of the data vectors and $r \in \{1,...,M\}\setminus y_i$ are multiclass labels excluding $y_i$.

Lee *et al.* (2001) and Schölkopf and Smola (2002) showed that the results from this method and the one-against-the-rest are similar. However, in this method, the optimization algorithm has to consider all the support vectors at the same time.

Therefore, it may be able to handle massive data sets but the memory requirement and thus, the computational time may be very high.

To summarize, it may be said that the choice of a multiclass method depends on the problem in hand. A user should consider the accuracy requirement, the computational time, the resources available and the nature of the problem. For example, the multiclass objective function approach may not be suitable for a problem that contains a large number of training samples and classes due to the requirement of large memory and extremely long computational time.

**3.5 Error-Correcting Output Code based approach**

The concept of Error- Correcting Output Coding (ECOC) based multi-class method is to apply binary (two-class) classifiers to solve the multi-class classification problems. This approach works by converting $M$ class classification problem into a large number $L$ of 2-class classification problems. ECOC assigns a unique code word to a class instead of assigning each class a label. A ($L$, $M$, $d$) error correcting code is a $L$ bit long, having $C$ unique code words with a Hamming distance of $d$. The hamming distance between two code words is the number of bit positions in which both differs. In a classification problem $M$ is the number of classes and $L$ is a number decided by the method used to generate error-correcting codes. Several methods such as Hadamard-Matrix codes, BCH codes (Bose and Ray-chauduri, 1960; Peterson and Weldon, 1972), random codes (James, 1998) and exhaustive codes (Dietterich and Bakiri, 1995) are proposed to generate error correcting codes. Dietterich and Bakiri, (1995) proposed to use codes with maximum Hamming distance between each other and suggested that it (d - 1)/2 errors can be corrected in the code words for a Hamming distance d between the codes.

Decomposition of a C class multi-class problem having $K_1,..........,K_C$ as the class labels generates a set of *m* binary classifiers represented by $f_1,.............f_m$. A binary classifier subdivides the input patterns into two complementary super classes $K_i^1$ and $K_i^{-1}$ grouping together one or more classes of multi-class problem. Let $M = [b_{ij}]$ is a decomposition matrix of dimension $m \times C$, connecting classes $K_1,..........,K_C$ to the super classes $K_i^1$ and $K_i^{-1}$, where an element of matrix M can be defined as:

$$b_{ij} = \begin{cases} 1 & if\ K_C \subseteq K_i^1 \\ -1 & if\ K_C \subseteq K_i^{-1} \end{cases}$$

Therefore, for *M* classes, a *coding matrix* $D \in \{\pm 1\}^{M \times C}$ is obtained.

When a new data is to be classified, the trained binary classifiers (or hypothesis) produce the estimated probability '$e_i$' that the test data comes from the *i*th super group one, thus producing a vector of probability estimates, $\mathbf{e} = (e_1, e_2,........,e_m)^T$ from all *p* binary classifiers. To classify the new data L1-distance for each of the *C* classes is calculated and the pixel is assigned to the new class with minimum L1-distance. The L1-distance is defined as:

$$L_i = \sum_{i=1}^{m} |e_i - o_{ij}|$$

Where $o_{ij}$ is the *j*th bit of the code word of the class *i*.

Allwein *et al.* (2000) proposed another scheme using a margin-based binary learning algorithm to replace the Hamming distance based decoding and proposed to use a *coding matrix* $D \in \{+1, 0, -1\}^{M \times C}$ in place of $\{\pm 1\}^{M \times C}$. This approach subdivides the input patterns into three complementary super classes $K_i^1$, $K_i^{-1}$ and some classes with zero as class label. During training process, a classification algorithm is provided with training data set labeled as $K_i^1$ and $K_i^{-1}$ while omitting all examples with class label zero.

Allwein *et al.* (2000) proposed and two types of random codes. The first type of random code called dense codes has $\lceil 10 \log_2 C \rceil$ columns for a problem with C classes. Dense random codes for each multiclass problem were chosen from $\{-1, 1\}$ by examining 10,000 random codes. These codes are chosen in a way to have largest hamming distance and have no identical column. The second type of code, called a sparse code, was chosen randomly from $\{-1, 0, 1\}$ having $\lceil 15 \log_2 K \rceil$ columns. Codes were selected by examining 10,000 random codes as in case of dense coding in way that no code had a row or column containing only zeros as well as to have maximum hamming distance. This study uses an exhaustive approach (Dietterich and Bakiri, 1995) as well as both approaches suggested by Allwein *et al.* (2000) to generate error correcting out put codes to solve a multiclass problem with support vector machine.

## 4. Data and Analysis

For this study, Landsat-7 Enhanced Thematic Mapper (ETM+) data (19/06/2000) of an agricultural area near Littleport (Cambridgeshire), UK was used. An area of 307-pixel (columns) by 330-pixel (rows) covering the area of interest was used for this study.

The classification problem involved the identification of seven land cover types (wheat, potato, sugar beet, onion, peas, lettuce and beans). Field Data printouts for the relevant crop seasons were collected from farmers and their representative agencies. The other areas were surveyed on the ground to prepare the ground reference image. A total of 4737 pixels were selected for all seven classes by using equalised random sampling. Pixels were then divided into two parts so as to remove any possible bias caused by using the same pixels for training and testing the classifiers. A total of 2700 training and 2037 test pixels were used. A radial basis kernel function with kernel width $\gamma = 2$ and regularisation parameter $C = 5000$ was used. All the processing with support vector machines was done on a window based Pentium IV processor with 256 MB of RAM was used.

## 5. Result and Conclusions

Table 1 provides the classification accuracy and training time with different multiclass approaches used in present study. The results suggest that except exhaustive technique based ECOC approach all multiclass methods provide comparable results in term of classification accuracy. An accuracy of 89% (kappa value = 0.87) is achieved with exhaustive technique based ECOC approach but at a large computation cost (806.6 minutes) in comparison to *one vs. one, one vs. rest, Weston and Watkins approach and DAG* approaches. A classification accuracy of 87.9% (kappa value = 0.86) is achieved by *one vs. one* with a very small training time of 6.4 seconds while *DAG approach* requires a training time of 6.5 seconds and achieves a classification accuracy of 87.63%.

Further, results using exhaustive technique based ECOC approach are not significantly better in comparison to *one vs. one* approach in term of classification accuracy. Approach suggested by Crammer and Singer (2001) requires a large training time (approx. 347 minutes) with no appreciable gain in term of classification accuracy in comparison to other multiclass approaches. Sparse random coding approach provides a classification accuracy of 87.19% in comparison to 85.32% with dense coding approach. This suggests a comparable performance by sparse coding approach to other multiclass approaches used in present study.

Table 1. Classification accuracies achieved with different multiclass approaches used in present study.

| Multiclass approach | Classification accuracy (%) |
|---|---|
| one against one | 87.90 |
| one against rest | 86.55 |
| Directed Acyclic Graph | 87.63 |
| Bound constrained approach | 87.29 |
| Crammer and Singer approach | 87.43 |
| ECOC (exhaustive approach) | 89.00 |
| ECOC (Dense coding approach) | 85.32 |
| ECOC (Sparse coding approach) | 87.19 |

Present study examined six approaches for the solution of multiclass classification problem using remote sensing data. *One against one* and *DAG* approach provide a comparable accuracy and requires almost same computational resources. The training time taken by *one against one and DAG* techniques is less than that with the *one against the rest* strategy. This study also concludes that the highest classification accuracy is achieved with exhaustive ECOC approach but requires very large training time. A comparison of accuracy achieved by exhaustive ECOC approach suggests no significant improvement in comparison to *one against one* approach. The main problem with the

'one against the rest' strategy is that it may produce unclassified data, and hence lower classification accuracies. Finally, results suggest the suitability of *One against one approach* for this type of data in term of classification accuracy and the computational cost. Further study is required to study the usefulness of this approach with other type of remote sensing data as well as data with large number classes.